\documentclass[10pt,twocolumn,letterpaper]{article}

\usepackage{cvpr,times,graphicx,tabulary,multirow,xspace,xcolor}
\usepackage{amsthm,amssymb,fixmath,mathtools,nicefrac}
\usepackage[pagebackref=true,breaklinks=true,letterpaper=true,colorlinks,
  citecolor=citecolor,bookmarks=false]{hyperref}
\definecolor{citecolor}{RGB}{34,139,34}

\usepackage[british,english,american]{babel}
\usepackage[title]{appendix}

\usepackage{subcaption}
\def\cvprPaperID{****}\cvprfinalcopy

\ifcvprfinal

\newcommand{\app}{\raise.17ex\hbox{$\scriptstyle\sim$}}

\newcommand{\dt}[1]{\fontsize{7pt}{0.1em}\selectfont (#1)}
\newcommand{\bd}[1]{\textbf{#1}}
\newcommand{\x}{\times}
\newcommand{\sqr}[1]{#1$^\textrm{2}$}
\newcommand{\aplvis}{AP$^\star$\xspace}

\newcommand{\name}{PointRend\xspace}

\newlength\savewidth\newcommand\shline{\noalign{\global\savewidth\arrayrulewidth
  \global\arrayrulewidth 1pt}\hline\noalign{\global\arrayrulewidth\savewidth}}
\newcolumntype{x}[1]{>{\centering\arraybackslash}p{#1pt}}
\newcommand{\tablestyle}[2]{\setlength{\tabcolsep}{#1}\renewcommand{\arraystretch}{#2}\centering\small}

\makeatletter
\renewcommand\paragraph{\@startsection{paragraph}{4}{\z@}%
  {.5em \@plus1ex \@minus.1ex}%
  {-.5em}%
  {\normalfont\normalsize\bfseries}}
\makeatother

\begin{document}
\title{\name: Image Segmentation as Rendering}
\author{%
 Alexander Kirillov \quad Yuxin Wu \quad Kaiming He \quad Ross Girshick\\[2mm]
 Facebook AI Research (FAIR)}
\maketitle

\begin{abstract}
We present a new method for efficient high-quality image segmentation of objects and scenes. By analogizing classical computer graphics methods for efficient rendering with over- and undersampling challenges faced in pixel labeling tasks, we develop a unique perspective of image segmentation as a rendering problem. From this vantage, we present the \emph{PointRend} (Point-based Rendering) neural network module: a module that performs point-based segmentation predictions at adaptively selected locations based on an iterative subdivision algorithm. PointRend can be flexibly applied to both instance and semantic segmentation tasks by building on top of existing state-of-the-art models. While many concrete implementations of the general idea are possible, we show that a simple design already achieves excellent results. Qualitatively, PointRend outputs crisp object boundaries in regions that are over-smoothed by previous methods. Quantitatively, PointRend yields significant gains on COCO and Cityscapes, for both instance and semantic segmentation. PointRend's efficiency enables output resolutions that are otherwise impractical in terms of memory or computation compared to existing approaches. Code has been made available at \url{https://github.com/facebookresearch/detectron2/tree/master/projects/PointRend}.
\end{abstract}

\section{Introduction}

Image segmentation tasks involve mapping pixels sampled on a regular grid to a label map, or a set of label maps, on the same grid. For semantic segmentation, the label map indicates the predicted category at each pixel. In the case of instance segmentation, a binary foreground \vs background map is predicted for each detected object. The modern tools of choice for these tasks are built on convolutional neural networks (CNNs) ~\cite{lecun1989backpropagation,Krizhevsky2012}.

\begin{figure}\centering\vspace{-0.2cm}
\includegraphics[width=1.0\linewidth]{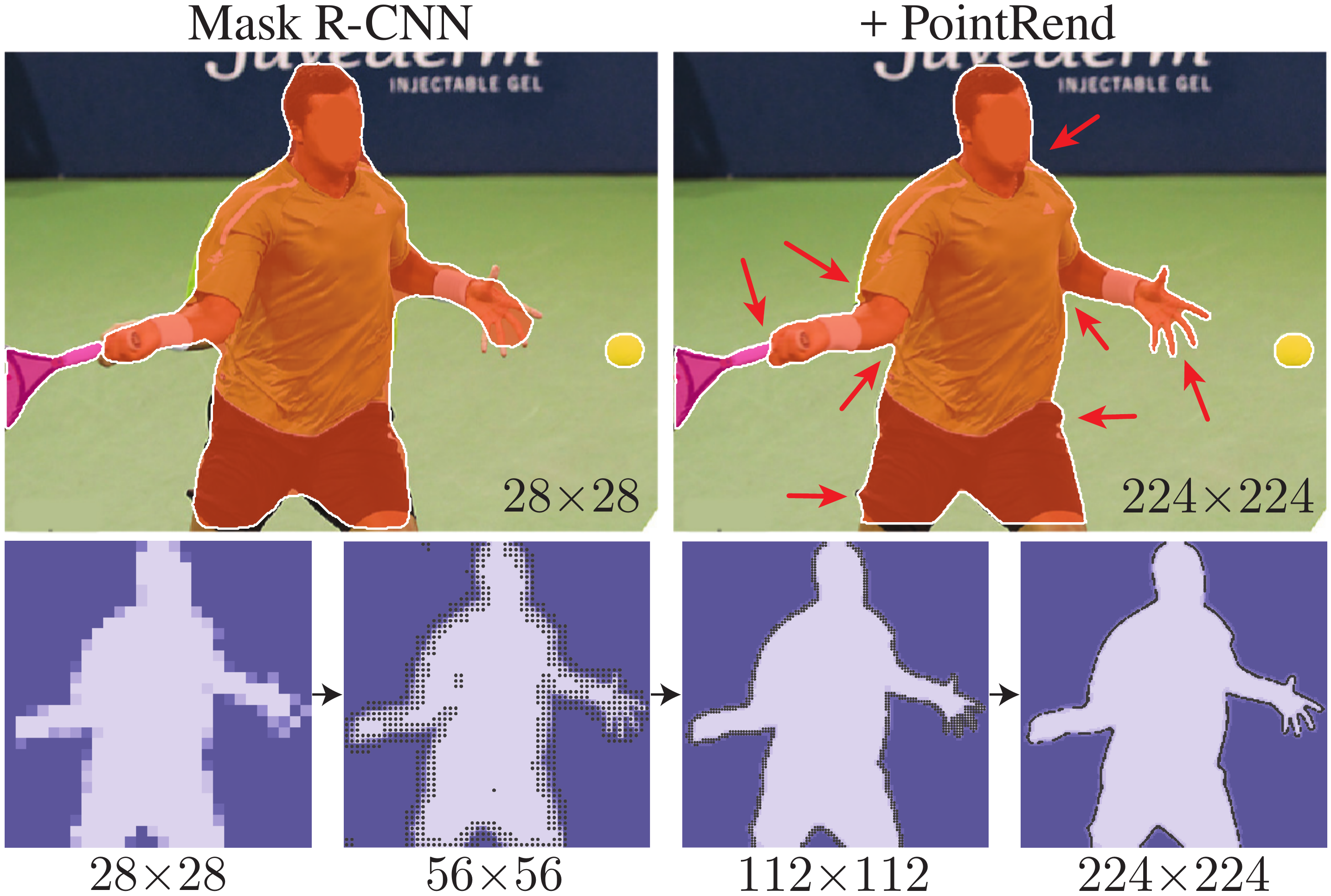}
\caption{\textbf{Instance segmentation with \name.} We introduce the \name (Point-based Rendering) module that makes predictions at adaptively sampled points on the image using a new point-based feature representation (see Fig.~\ref{fig:architecture}). \name is general and can be flexibly integrated into existing semantic and instance segmentation systems. When used to replace Mask R-CNN's default mask head~\cite{he2017mask} (top-left), PointRend yields significantly more detailed results (top-right). (bottom) During inference, \name iterative computes its prediction. Each step applies bilinear upsampling in smooth regions and makes higher resolution predictions at a small number of adaptively selected points that are likely to lie on object boundaries (black points). All figures in the paper are best viewed digitally with zoom. Image source:~\cite{paphio}.}
\label{fig:teaser}
\end{figure}

\begin{figure*}
\includegraphics[width=1.0\linewidth]{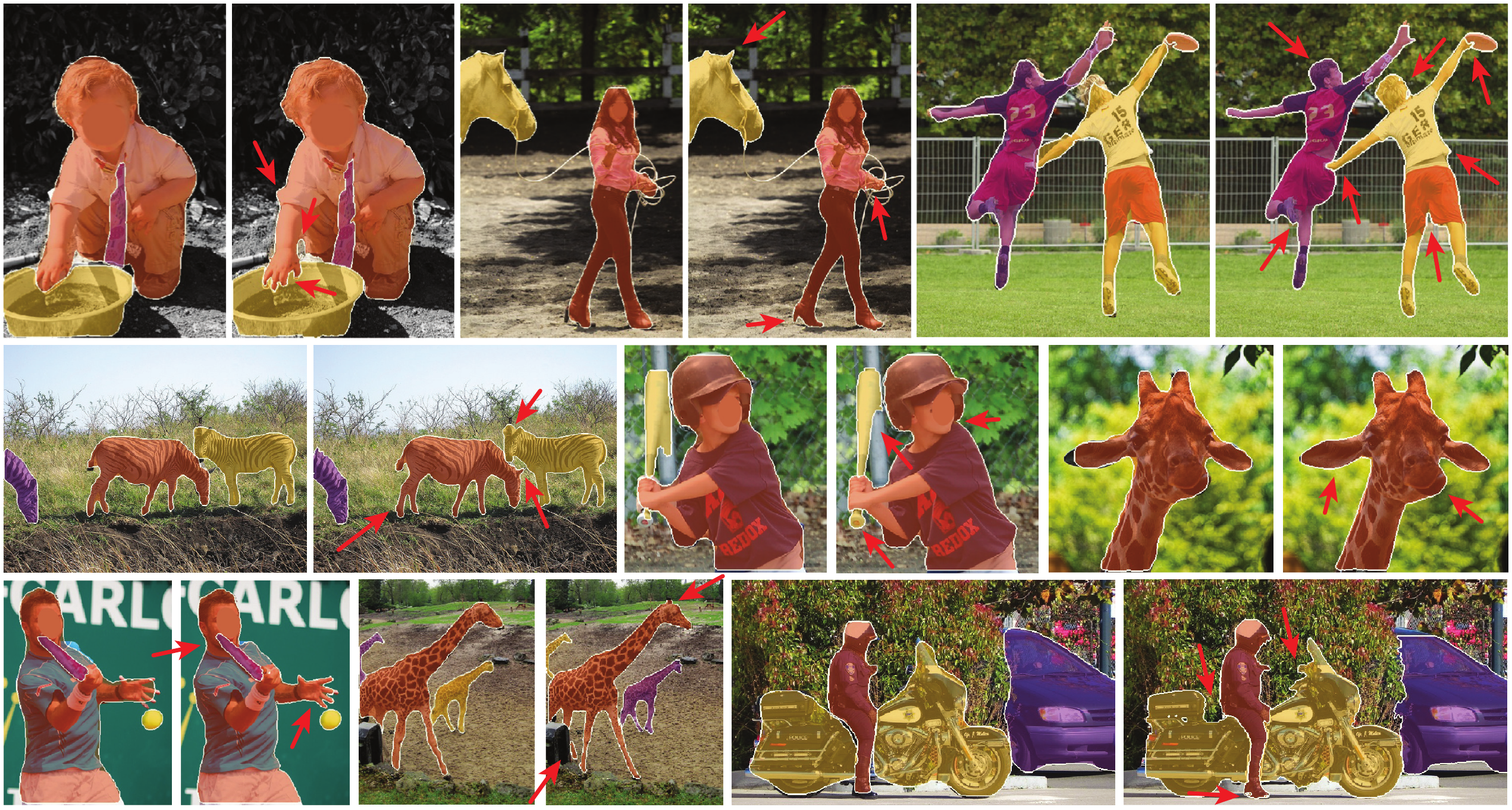}
  \caption{\textbf{Example result pairs from Mask R-CNN~\cite{he2017mask} with its standard mask head (left image) \vs with \name (right image)}, using ResNet-50~\cite{he2016deep} with FPN~\cite{lin2016feature}. Note how \name predicts masks with substantially finer detail around object boundaries.}
\vspace{-2mm}
\label{fig:examples}
\end{figure*}

CNNs for image segmentation typically operate on regular grids: the input image
is a regular grid of pixels, their hidden representations are feature vectors on
a regular grid, and their outputs are label maps on a regular grid. Regular
grids are convenient, but not necessarily computationally ideal for image
segmentation. The label maps predicted by these networks should be mostly
smooth, \ie, neighboring pixels often take the same label, because
high-frequency regions are restricted to the sparse boundaries between
objects. A regular grid will unnecessarily oversample the smooth areas while
simultaneously undersampling object boundaries. The result is excess
computation in smooth regions and blurry contours (Fig.~\ref{fig:teaser},
upper-left).
Image segmentation methods often predict labels on a low-resolution regular
grid, \eg, 1/8-th of the input \cite{long2015fully} for semantic segmentation, or 28$\x$28 \cite{he2017mask} for instance segmentation,
as a compromise between undersampling and oversampling.

Analogous sampling issues have been studied for decades in computer graphics. For example, a \emph{renderer} maps a model (\eg, a 3D mesh) to a rasterized image, \ie a regular grid of pixels. While the output is on a regular grid, computation is not allocated uniformly over the grid. Instead, a common graphics strategy is to compute pixel values at an \emph{irregular} subset of adaptively selected \emph{points} in the image plane. The classical \emph{subdivision} technique of~\cite{whitted1979improved}, as an example, yields a quadtree-like sampling pattern that efficiently renders an anti-aliased, high-resolution image.

The central idea of this paper is to view image segmentation as a rendering problem and to adapt classical ideas from computer graphics to efficiently ``render'' high-quality label maps (see Fig.~\ref{fig:teaser}, bottom-left). We encapsulate this computational idea in a new neural network module, called \bd{\name}, that uses a subdivision strategy to adaptively select a non-uniform set of points at which to compute labels. \name can be incorporated into popular meta-architectures for both instance segmentation (\eg, Mask R-CNN~\cite{he2017mask}) and semantic segmentation (\eg, FCN~\cite{long2015fully}). Its subdivision strategy efficiently computes high-resolution segmentation maps using an order of magnitude fewer floating-point operations than direct, dense computation.

\name is a general module that admits many possible implementations. Viewed 
abstractly, a \name module accepts one or more typical CNN feature maps
$f(x_{i}, y_{i})$ that are
defined over regular grids, and outputs
high-resolution predictions $p(x'_i,y'_i)$ over a finer grid.
Instead of making excessive predictions over all points on the output grid,
\name makes predictions only on carefully selected points.
To make these predictions, it extracts a point-wise feature representation for the selected points
by interpolating $f$, and uses a small \emph{point head} subnetwork to predict output labels from
the point-wise features.
We will present a simple and effective \name implementation.

We evaluate \name on instance and semantic segmentation tasks using the COCO~\cite{lin2014coco} and Cityscapes~\cite{Cordts2016Cityscapes} benchmarks. Qualitatively, \name efficiently computes sharp boundaries between objects, as illustrated in Fig.~\ref{fig:examples} and Fig.~\ref{fig:examples-cs}. We also observe quantitative improvements even though the standard intersection-over-union based metrics for these tasks (mask AP and mIoU) are biased towards object-interior pixels and are relatively insensitive to boundary improvements. \name improves strong Mask R-CNN and DeepLabV3~\cite{deeplabV3} models by a significant margin.

\section{Related Work}

\paragraph{Rendering} algorithms in computer graphics output a regular grid of pixels. However, they usually compute these pixel values over a non-uniform set of points. Efficient procedures like subdivision~\cite{whitted1979improved} and adaptive sampling~\cite{mitchell1987generating,pbrtchapter7} refine a coarse rasterization in areas where pixel values have larger variance. Ray-tracing renderers often use oversampling~\cite{zhou2008real}, a technique that samples some points more densely than the output grid to avoid aliasing effects. Here, we apply classical subdivision to image segmentation.

\paragraph{Non-uniform grid representations.} Computation on regular grids is the dominant paradigm for 2D image analysis, but this is not the case for other vision tasks. In 3D shape recognition, large 3D grids are infeasible due to cubic scaling. Most CNN-based approaches do not go beyond coarse 64$\x$64$\x$64 grids~\cite{girdhar2016learning,choy20163d}. Instead, recent works consider more efficient non-uniform representations such as meshes~\cite{wang2018pixel2mesh,gkioxari2019mesh}, signed distance functions~\cite{mescheder2019occupancy}, and octrees~\cite{tatarchenko2017octree}. Similar to a signed distance function, \name can compute segmentation values at any point.

Recently, Marin \etal~\cite{marin2019efficient} propose an efficient semantic segmentation network based on non-uniform subsampling of the \emph{input} image prior to processing with a standard semantic segmentation network. \name, in contrast, focuses on non-uniform sampling at the \emph{output}. It may be possible to combine the two approaches, though~\cite{marin2019efficient} is currently unproven for instance segmentation.

\paragraph{Instance segmentation} methods based on the Mask R-CNN
meta-architecture~\cite{he2017mask} occupy top ranks in recent
challenges~\cite{liu2018path,chen2019hybrid}. These region-based architectures
typically predict masks on a 28$\x$28 grid irrespective of object size. This is
sufficient for small objects, but for large objects it produces
undesirable ``blobby'' output that over-smooths the fine-level details of large
objects (see Fig.~\ref{fig:teaser}, top-left). Alternative, bottom-up approaches
group pixels to form object
masks~\cite{liu2017sgn,arnab2017pixelwise,kirillov2016instancecut}. These
methods can produce more detailed output, however, they lag behind region-based
approaches on most instance segmentation
benchmarks~\cite{lin2014coco,Cordts2016Cityscapes,neuhold2017mapillary}.
TensorMask~\cite{chen2019tensormask}, an alternative sliding-window method, uses
a sophisticated network design to predict sharp high-resolution masks for large
objects, but its accuracy also lags slightly behind.
In this paper, we show that a region-based segmentation model equipped with
\name can produce masks with fine-level details while improving the accuracy of region-based approaches.

\paragraph{Semantic segmentation.} Fully convolutional networks (FCNs)~\cite{long2015fully} are the foundation of modern semantic segmentation approaches. They often predict outputs that have lower resolution than the input grid and use bilinear upsampling to recover the remaining 8-16$\x$ resolution. Results may be improved with dilated/atrous convolutions that replace some subsampling layers~\cite{deeplabV2,deeplabV3} at the expense of more memory and computation.

Alternative approaches include encoder-decoder achitectures~\cite{deeplabV3plus,kirillov2019panopticfpn,ronneberger2015u,sun2019high} that subsample the grid representation in the encoder and then upsample it in the decoder, using skip connections~\cite{ronneberger2015u} to recover filtered details. Current approaches combine dilated convolutions with an encoder-decoder structure~\cite{deeplabV3plus,liu2019auto} to produce output on a 4$\x$ sparser grid than the input grid before applying bilinear interpolation. In our work, we propose a method that can efficiently predict fine-level details on a grid as dense as the input grid.

\begin{figure}\centering
\includegraphics[width=1.0\linewidth]{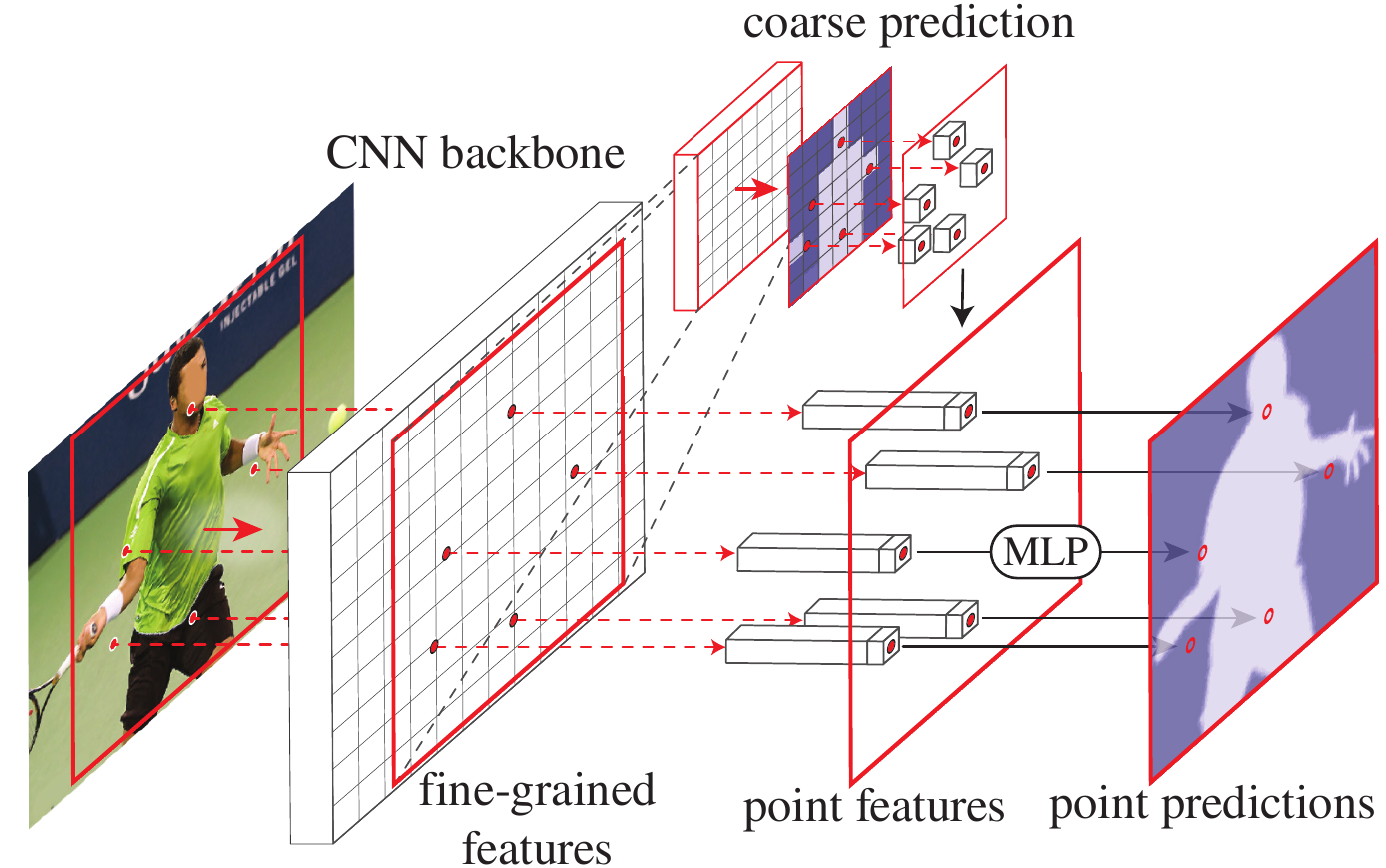}
  \caption{\textbf{\name applied to instance segmentation}. A standard network for instance segmentation
    (solid red arrows) takes an input image and yields a coarse (\eg 7$\x$7)
    mask prediction for each detected object (red box) using a lightweight
    segmentation head. To refine the coarse mask, \name selects a set of points
    (red dots) and makes prediction for each point independently with a small
    MLP\@. The MLP uses interpolated features computed at these points 
    (dashed red arrows) from (1) a fine-grained feature map of the backbone CNN
    and (2) from the coarse
    prediction mask. The coarse mask
    features enable the MLP to make different predictions at a single point that
    is contained by two or more boxes. The proposed subdivision mask rendering
    algorithm (see Fig.~\ref{fig:subdivision} and \S\ref{subsec:sampling}) applies this process iteratively to refine uncertain regions of the predicted mask.}
\label{fig:architecture}
\end{figure}

\section{Method}

We analogize image segmentation (of objects and/or scenes) in computer vision to
image rendering in computer graphics. Rendering is about displaying a model
(\eg, a 3D mesh) as a regular grid of pixels, \ie, an image. While the output
representation is a regular grid, the underlying physical entity (\eg, the 3D
model) is continuous and its physical occupancy and other attributes can be
queried at any \emph{real-value point} on the image plane using physical and geometric reasoning, such as ray-tracing.

Analogously, in computer vision, we can think of an image segmentation as the occupancy map of an underlying continuous entity, and the segmentation output, which is a regular grid of predicted labels, is ``rendered" from it. The entity is encoded in the network's feature maps and can be accessed at any point by interpolation. A parameterized function, that is trained to predict occupancy from these interpolated point-wise feature representations, is the counterpart to physical and geometric reasoning.

Based on this analogy, we propose \name (\emph{Point-based Rendering}) as a
methodology for image segmentation using point representations.
A \name module accepts one or more typical CNN feature maps of $C$ channels $f \in
\mathbb{R}^{C\times H\times W}$, each defined over a regular grid (that is typically 4$\x$ to 16$\x$ coarser than the image grid), and outputs
predictions for the $K$ class labels $p \in \mathbb{R}^{K\times H' \times W'}$ over a regular grid of
different (and likely higher) resolution.
A \name module consists of three main components:
(i) A \emph{point selection strategy} chooses a small number of real-value
points to make predictions on, avoiding excessive computation for all pixels in the
high-resolution output grid.
(ii) For each selected point, a \emph{point-wise feature representation} is extracted.
Features for a real-value
point are computed by bilinear interpolation of $f$, using
the point's 4 nearest neighbors that are on the regular grid of $f$.
As a result, it is able to utilize sub-pixel information encoded
in the channel dimension of $f$ to predict a segmentation that has higher resolution
than $f$.
(iii) A \emph{point head}: a small neural network trained to predict a label from this point-wise
feature representation, independently for each point.

The \name architecture can be applied to instance segmentation (\eg, on Mask
R-CNN~\cite{he2017mask}) and semantic segmentation (\eg, on
FCNs~\cite{long2015fully}) tasks. For instance segmentation, \name is applied to
each region. It computes masks in a coarse-to-fine fashion by
making predictions over a set of selected points (see
Fig.~\ref{fig:architecture}).
For semantic segmentation, the whole image can be considered as a single region, and thus without loss of generality we will describe \name in the context of instance segmentation.
We discuss the three main components in more detail next.

\subsection{Point Selection for Inference and Training}
\label{subsec:sampling}

At the core of our method is the idea of flexibly and adaptively selecting points in the image plane at which to predict segmentation labels. Intuitively, these points should be located more densely near high-frequency areas, such as object boundaries, analogous to the anti-aliasing problem in ray-tracing. We develop this idea for inference and training.

\paragraph{Inference.} Our selection strategy for inference is inspired by the classical technique of \emph{adaptive subdivision} \cite{whitted1979improved} in computer graphics. The technique is used to efficiently render high resolutions images (\eg, via ray-tracing) by computing only at locations where there is a high chance that the value is significantly different from its neighbors; for all other locations the values are obtained by interpolating already computed output values (starting from a coarse grid).

\begin{figure}\centering
\includegraphics[width=1.0\linewidth,height=0.286\linewidth]{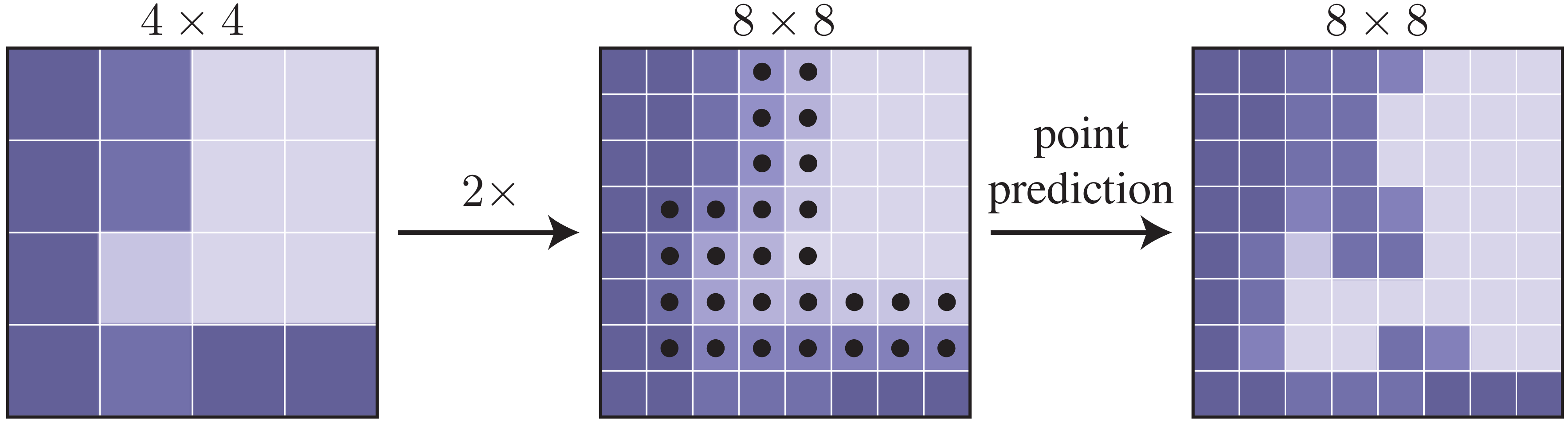}
  \caption{\textbf{Example of one adaptive subdivision step}. A prediction on a 4$\x$4 grid is upsampled by 2$\x$ using bilinear interpolation. Then, \name makes prediction for the $N$ most ambiguous points (black dots) to recover detail on the finer grid. This process is repeated until the desired grid resolution is achieved.}
\label{fig:subdivision}
\end{figure}

For each region, we iteratively ``render" the output mask in a coarse-to-fine fashion. The coarsest level prediction is made on the points on a regular grid (\eg, by using a standard coarse segmentation prediction head). In each iteration, \name upsamples its previously predicted segmentation using bilinear interpolation and then selects the $N$ most uncertain points (\eg, those with probabilities closest to 0.5 for a binary mask) on this denser grid. \name then computes the point-wise feature representation (described shortly in \S\ref{subsec:representations}) for each of these $N$ points and predicts their labels. This process is repeated until the segmentation is upsampled to a desired resolution. One step of this procedure is illustrated on a toy example in Fig.~\ref{fig:subdivision}.

With a desired output resolution of $M$$\x$$M$ pixels and a starting resolution of $M_0$$\x$$M_0$, \name requires no more than $N\log_2\frac{M}{M_0}$ point predictions. This is much smaller than $M$$\x$$M$, allowing \name to make high-resolution predictions much more effectively. For example, if $M_0$ is 7 and the desired resolutions is $M{=}224$, then 5 subdivision steps are preformed. If we select $N{=}28^2$ points at each step, \name makes predictions for only \sqr{28}$\cdot$4.25 points, which is 15 times smaller than \sqr{224}. Note that fewer than $N\log_2\frac{M}{M_0}$ points are selected overall because in the first subdivision step only \sqr{14} points are available.

\begin{figure}\centering
\includegraphics[width=1.0\linewidth,height=0.33\linewidth]{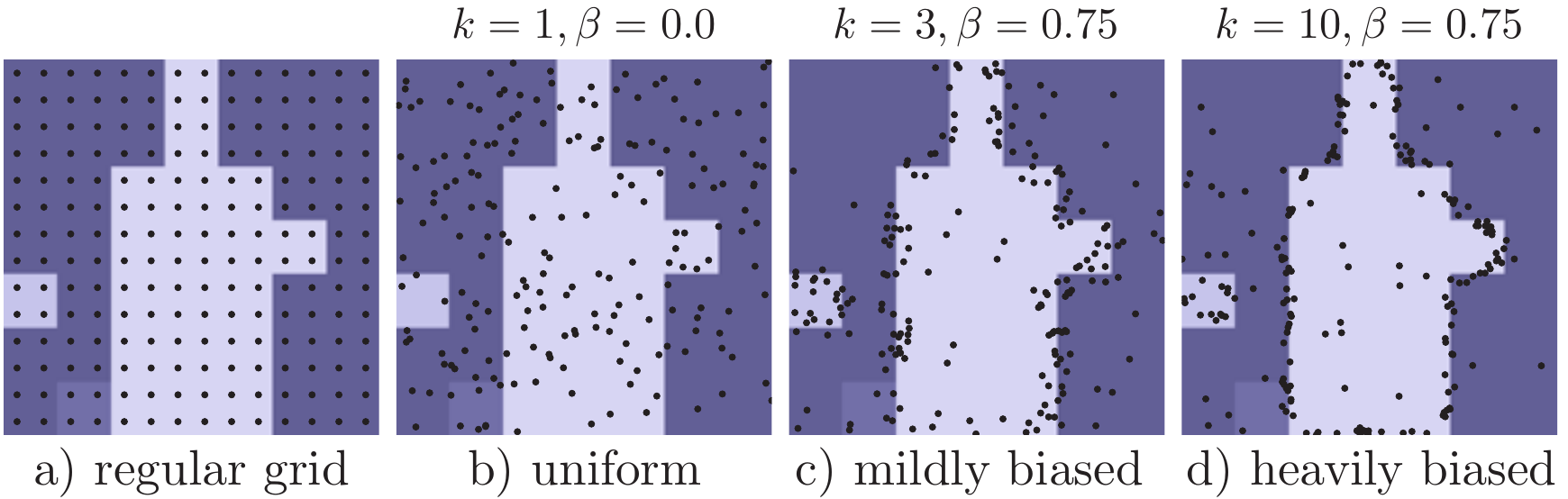}
  \caption{\textbf{Point sampling during training.} We show $N{=}14^2$ points sampled using different strategies for the same underlying coarse prediction. To achieve high performance only a small number of points are sampled per region with a mildly biased sampling strategy making the system more efficient during training.}
\label{fig:sampling}
\end{figure}

\paragraph{Training.} During training, \name also needs to select points
at which to construct point-wise features for training the point head.
In principle, the point selection strategy can be similar to the subdivision
strategy used in inference. However, subdivision introduces sequential steps
that are less friendly to training neural networks with backpropagation.
Instead, for training we use a non-iterative strategy based on random sampling.

The sampling strategy selects $N$ points on a feature map to train on.\footnote{The value of $N$ can be different for training and inference selection.}
It is designed to bias selection towards uncertain regions, while also retaining some degree of uniform coverage, using three principles. \mbox{(i) \emph{Over generation}:} we over-generate candidate points by randomly sampling $kN$ points ($k{>}1$) from a uniform distribution. \mbox{(ii) \emph{Importance sampling}:} we focus on points with uncertain coarse predictions by interpolating the coarse prediction values at all $kN$ points and computing a task-specific uncertainty estimate (defined in \S\ref{sec:instance} and \S\ref{sec:semantic}). The most uncertain $\beta N$ points ($\beta\in[0,1]$) are selected from the $kN$ candidates. \mbox{(iii) \emph{Coverage}:} the remaining $(1-\beta)N$ points are sampled from a  uniform distribution.  We illustrate this procedure with different settings, and compare it to regular grid selection, in Fig.~\ref{fig:sampling}.

At training time, predictions and loss functions are only computed on the $N$ sampled
points (in addition to the coarse segmentation),
which is simpler and more efficient than backpropagation through subdivision steps. This design is similar to the parallel training of \mbox{RPN~+~Fast~R-CNN} in a Faster R-CNN system~\cite{girshick2015fast}, whose inference is sequential.

\subsection{Point-wise Representation and Point Head}
\label{subsec:representations}

\name constructs \emph{point-wise features} at selected points by combining (\eg, concatenating) two feature types, fine-grained and coarse prediction features, described next.

\paragraph{Fine-grained features.} To allow \name to render fine segmentation details
we extract a feature vector at each sampled point from CNN feature maps. Because
a point is a real-value 2D coordinate, we perform bilinear interpolation on the feature maps to compute the feature vector, following standard practice \cite{jaderberg2015spatial,he2017mask,dai2017deformable}. Features can be extracted from a single feature map (\eg, res$_2$ in a ResNet); they can also be extracted from multiple feature maps (\eg, res$_2$ to res$_5$, or their feature pyramid \cite{lin2016feature} counterparts) and concatenated, following the Hypercolumn method \cite{hariharan2015hypercolumns}.

\paragraph{Coarse prediction features.} 
The fine-grained features enable resolving detail, but are also deficient in two regards.
First, they do not contain region-specific information and thus the same point
overlapped by two instances' bounding boxes will have the same fine-grained features. Yet,
the point can only be in the foreground of one instance. Therefore, for the task of
instance segmentation, where different regions may predict different labels for the same
point, additional region-specific information is needed.

Second, depending on which feature maps are used for the fine-grained features, the features may contain only relatively low-level information (\eg, we will use res$_2$ with DeepLabV3). In this case, a feature source with more contextual and semantic information can be helpful. This issue affects both instance and semantic segmentation.

Based on these considerations, the second feature type is a coarse segmentation prediction from the network, \ie, a $K$-dimensional vector at each point in the region (box) representing a $K$-class prediction. The coarse resolution, by design, provides more globalized context, while the channels convey the semantic classes.
These coarse predictions are similar to the outputs made by the existing
architectures, and are supervised during training in the same way as existing
models.
For instance segmentation, the coarse prediction can be, for example, the output of a lightweight 7$\x$7 resolution mask head in Mask R-CNN. For semantic segmentation, it can be, for example, predictions from a stride 16 feature map.

\paragraph{Point head.} Given the point-wise feature representation at each selected point, \name makes point-wise segmentation predictions using a simple multi-layer perceptron (MLP). This MLP shares weights across all points (and all regions), analogous to a graph convolution \cite{kipf2016semi} or a \mbox{PointNet} \cite{qi2017pointnet}.
Since the MLP predicts a segmentation label for each point, it can be trained by standard
task-specific segmentation losses (described in \S\ref{sec:instance} and \S\ref{sec:semantic}).

\section{Experiments: Instance Segmentation}\label{sec:instance}

\paragraph{Datasets.} We use two standard instance segmentation datasets: COCO~\cite{lin2014coco} and Cityscapes~\cite{Cordts2016Cityscapes}. We report the standard mask AP metric~\cite{lin2014coco} using  the median of 3 runs for COCO and 5 for Cityscapes (it has higher variance).

COCO has 80 categories with instance-level annotation. We train on \texttt{train2017} ($\app$118k images) and report results on \texttt{val2017} (5k images). As noted in~\cite{gupta2019lvis}, the COCO ground-truth is often coarse and AP for the dataset may not fully reflect improvements in mask quality. Therefore we supplement COCO results with AP measured using the 80 COCO category subset of LVIS~\cite{gupta2019lvis}, denoted by \aplvis. The LVIS annotations have significantly higher quality. Note that for \aplvis we use the same models trained on COCO and simply re-evaluate their predictions against the higher-quality LVIS annotations using the LVIS evaluation API.

Cityscapes is an ego-centric street-scene dataset with 8 categories, 2975 train images, and 500 validation images. The images are higher resolution compared to COCO (1024$\x$2048 pixels) and have finer, more pixel-accurate ground-truth instance segmentations.

\paragraph{Architecture.} Our experiments use Mask R-CNN with a ResNet-50~\cite{he2016deep} + FPN~\cite{lin2016feature} backbone. The default mask head in Mask R-CNN is a region-wise FCN, which we denote by ``4$\x$ conv''.\footnote{Four layers of 3$\x$3 convolutions with 256 output channels are applied to a 14$\x$14 input feature map. Deconvolution with a 2$\x$2 kernel transforms this to 28$\x$28. Finally, a 1$\x$1 convolution predicts mask logits.} We use this as our baseline for comparison. For \name, we make appropriate modifications to this baseline, as described next.

\paragraph{Lightweight, coarse mask prediction head.} To compute the coarse
prediction, we replace the 4$\x$ conv mask head with a lighter weight
design that resembles Mask R-CNN's box head and produces a 7$\x$7 mask prediction.
Specifically, for each bounding box, we extract a 14$\x$14 feature map from the
P$_2$ level of the FPN using bilinear interpolation. The features are computed on a regular grid inside the bounding box (this operation can seen as a simple version of RoIAlign). Next, we use a stride-two 2$\x$2 convolution layer with 256 output channels followed by ReLU~\cite{nair2010rectified}, which reduces the spatial size to 7$\x$7. Finally, similar to Mask R-CNN's box head, an MLP with two 1024-wide hidden layers is applied to yield a 7$\x$7 mask prediction for each of the $K$ classes. ReLU is used on the MLP's hidden layers and the sigmoid activation function is applied to its outputs.

\paragraph{\name.} At each selected point, a $K$-dimensional feature vector is extracted from the coarse prediction head's output using bilinear interpolation. \name also interpolates a 256-dimensional feature vector from the P$_2$ level of the FPN. This level has a stride of 4 \wrt the input image. These coarse prediction and fine-grained feature vectors are concatenated. We make a $K$-class prediction at selected points using an MLP with 3 hidden layers with 256 channels. In each layer of the MLP, we supplement the 256 output channels with the $K$ coarse prediction features to make the input vector for the next layer. We use ReLU inside the MLP and apply sigmoid to its output.

\paragraph{Training.} We use the standard 1$\x$ training schedule and data augmentation from Detectron2~\cite{wu2019detectron2} by default (full  details are in the appendix).
For \name, we sample \sqr{14} points using the biased sampling strategy described in the \S\ref{subsec:sampling} with $k{=}3$ and $\beta{=}0.75$. We use the distance between 0.5 and the probability of the ground truth class interpolated from the coarse prediction as the point-wise uncertainty measure. For a predicted box with ground-truth class $c$, we sum the binary cross-entropy loss for the $c$-th MLP output over the \sqr{14} points. The lightweight coarse prediction head uses the average cross-entropy loss for the mask predicted for class $c$, \ie, the same loss as the baseline 4$\x$ conv head. We sum all losses without any re-weighting.

During training, Mask R-CNN applies the box and mask heads in parallel, while during inference they run as a cascade. We found that training as a cascade does not improve the baseline Mask R-CNN, but \name can benefit from it by sampling points inside more accurate boxes, slightly improving overall performance ($\app$0.2\% AP, absolute).

\paragraph{Inference.} For inference on a box with predicted class $c$, unless otherwise specified, we use the adaptive subdivision technique to refine the coarse 7$\x$7 prediction for class $c$ to the 224$\x$224 in 5 steps. At each step, we select and update (at most) the $N{=}28^2$ most uncertain points based on the absolute difference between the predictions and 0.5.

\subsection{Main Results}

\begin{table}[t]\centering
\tablestyle{5pt}{1.0}\begin{tabular}{@{}l|c|cc|c@{}}
& output & \multicolumn{2}{c|}{COCO} & Cityscapes \\
mask head & resolution & AP & \aplvis & AP \\\shline
  4$\x$ conv                           & 28$\x$28    & 35.2 \phantom{\dt{+1.1}} & 37.6 \phantom{\dt{+2.1}} & 33.0 \phantom{\dt{+2.8}}\\
PointRend                            & 28$\x$28    & 36.1 \dt{+0.9} & 39.2 \dt{+1.6} & 35.5 \dt{+2.5}\\
PointRend                            & 224$\x$224  & \bd{36.3 \dt{+1.1}} & \bd{39.7 \dt{+2.1}} & \bd{35.8 \dt{+2.8}}
\end{tabular}
  \vspace{-1mm}
  \caption{\bd{PointRend \vs the default 4$\x$ conv mask head for Mask R-CNN~\cite{he2017mask}.} Mask AP is reported. \aplvis is COCO mask AP evaluated against the higher-quality LVIS annotations~\cite{gupta2019lvis} (see text for details). A ResNet-50-FPN backbone is used for both COCO and Cityscapes models. PointRend outperforms the standard 4$\x$ conv mask head both quantitively and qualitatively. Higher output resolution leads to more detailed predictions, see Fig.~\ref{fig:examples} and Fig.~\ref{fig:coarse-vs-fine}.
\label{tab:instance-main}}
  \vspace{-2mm}
\end{table}

\begin{figure}\centering
\includegraphics[width=1.0\linewidth]{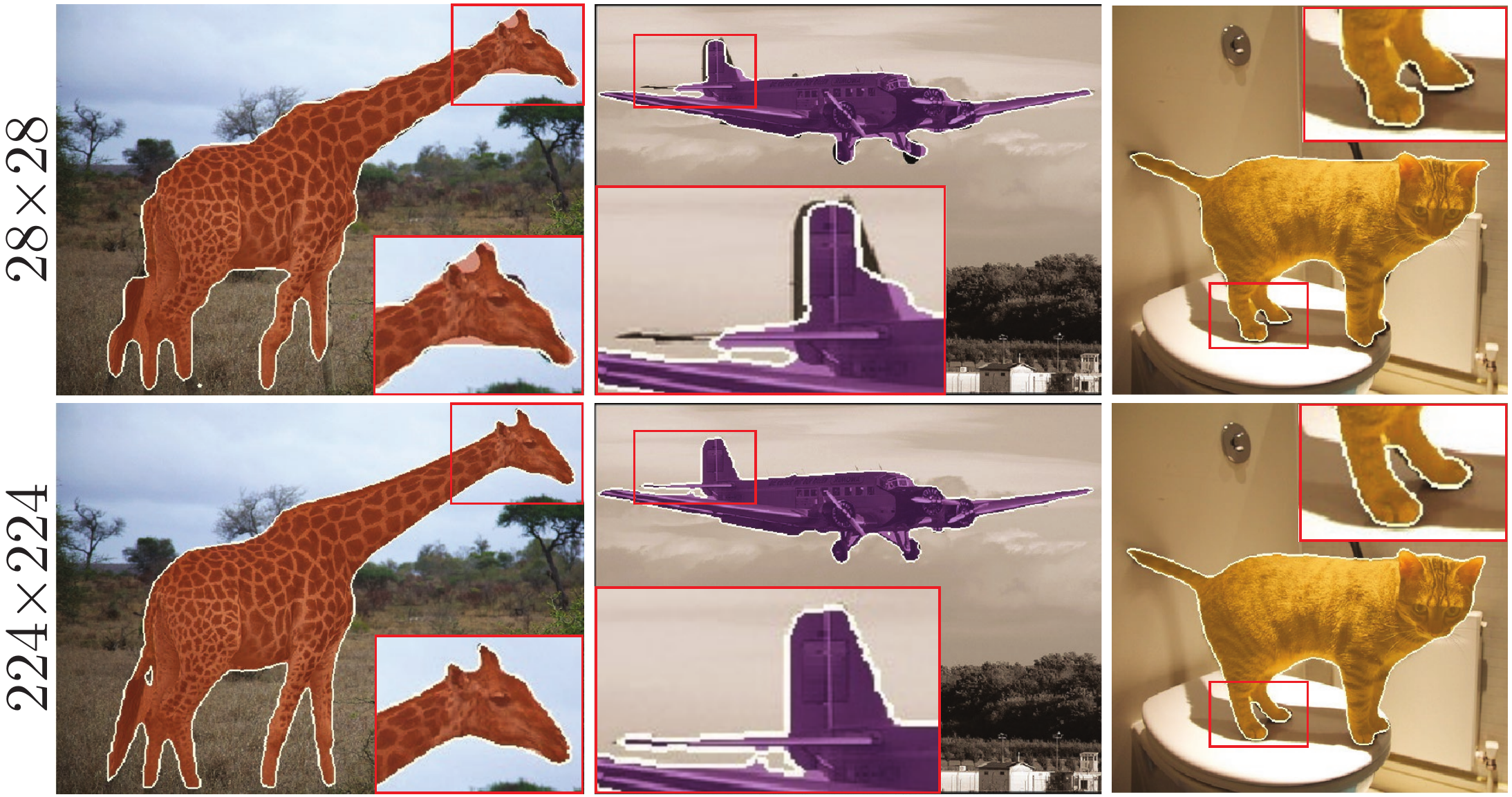}
  \caption{\textbf{\name inference with different output resolutions}. High resolution masks align better with object boundaries.}
\label{fig:coarse-vs-fine}
\end{figure}

We compare \name to the default 4$\x$ conv head in Mask R-CNN in Table~\ref{tab:instance-main}. \name outperforms the default head on both datasets. The gap is larger when evaluating the COCO categories using the LVIS annotations (\aplvis) and for Cityscapes, which we attribute to the superior annotation quality in these datasets. Even with the same output resolution \name outperforms the baseline. The difference between 28$\x$28 and 224$\x$224 is relatively small because AP uses intersection-over-union~\cite{everingham2015pascal} and, therefore, is heavily biased towards object-interior pixels and less sensitive to the boundary quality. Visually, however, the difference in boundary quality is obvious, see Fig.~\ref{fig:coarse-vs-fine}.

\begin{table}[t]\centering
\tablestyle{5pt}{1.0}\begin{tabular}{@{}l|c|cc@{}}
mask head                   & output resolution & FLOPs & \# activations \\\shline
4$\x$ conv                  & 28$\x$28    & 0.5B & 0.5M \\
4$\x$ conv                  & 224$\x$224  & 34B & 33M \\
PointRend                   & 224$\x$224  & 0.9B & 0.7M \\
\end{tabular}
  \vspace{-1mm}
\caption{\bd{FLOPs (multiply-adds) and activation counts for a 224$\x$224 output resolution mask.} \name's efficient subdivision makes 224$\x$224 output feasible in contrast to the standard 4$\x$ conv mask head modified to use an RoIAlign size of 112$\x$112.
\label{tab:compute}}
  \vspace{-6mm}
\end{table}

\paragraph{Subdivision inference} allows \name to yield a high resolution 224$\x$224 prediction using more than 30 times less compute (FLOPs) and memory than the default 4$\x$ conv head needs to output the same resolution (based on taking a 112$\x$112 RoIAlign input), see Table~\ref{tab:compute}. \name makes high resolution output feasible in the Mask R-CNN framework by ignoring areas of an object where a coarse prediction is sufficient (\eg, in the areas far away from object boundaries). In terms of wall-clock runtime, our \emph{unoptimized} implementation outputs 224$\x$224 masks at $\app$13 fps, which is roughly the same frame-rate as a 4$\x$ conv head modified to output 56$\x$56 masks (by doubling the default RoIAlign size), a design that actually has \emph{lower} COCO AP compared to the 28$\x$28 4$\x$ conv head (34.5\% \vs. 35.2\%).

\begin{table}[t]\centering
  \tablestyle{5pt}{1.0}\begin{tabular}{@{}c|c|cc|c@{}}
    & \# points per & \multicolumn{2}{c|}{COCO} & Cityscapes \\
    output resolution & subdivision step & AP & \aplvis & AP \\\shline
28$\x$28          & \sqr{28}  & 36.1 & 39.2 & 35.4 \\
56$\x$56          & \sqr{28}  & 36.2 & 39.6 & \underline{35.8} \\
112$\x$112        & \sqr{28}  & \underline{36.3} & \underline{39.7} & 35.8 \\
224$\x$224        & \sqr{28}  & 36.3 & 39.7 & 35.8 \\\hline
224$\x$224        & \sqr{14}  & 36.1 & 39.4 & 35.5 \\
224$\x$224        & \sqr{28}  & \underline{36.3} & \underline{39.7} & \underline{35.8} \\
224$\x$224        & \sqr{56}  & 36.3 & 39.7 & 35.8 \\
224$\x$224        & \sqr{112} & 36.3 & 39.7 & 35.8 \\
\end{tabular}
\vspace{-1mm}
\caption{\bd{Subdivision inference parameters.} Higher output resolution improves AP. Although improvements saturate quickly (at underlined values) with the number of points sampled at each subdivision step, qualitative results may continue to improve for complex objects. \aplvis is COCO mask AP evaluated against the higher-quality LVIS annotations~\cite{gupta2019lvis} (see text for details).
\label{tab:subdivision}}
\end{table}

\begin{figure}\centering
\includegraphics[width=1.0\linewidth]{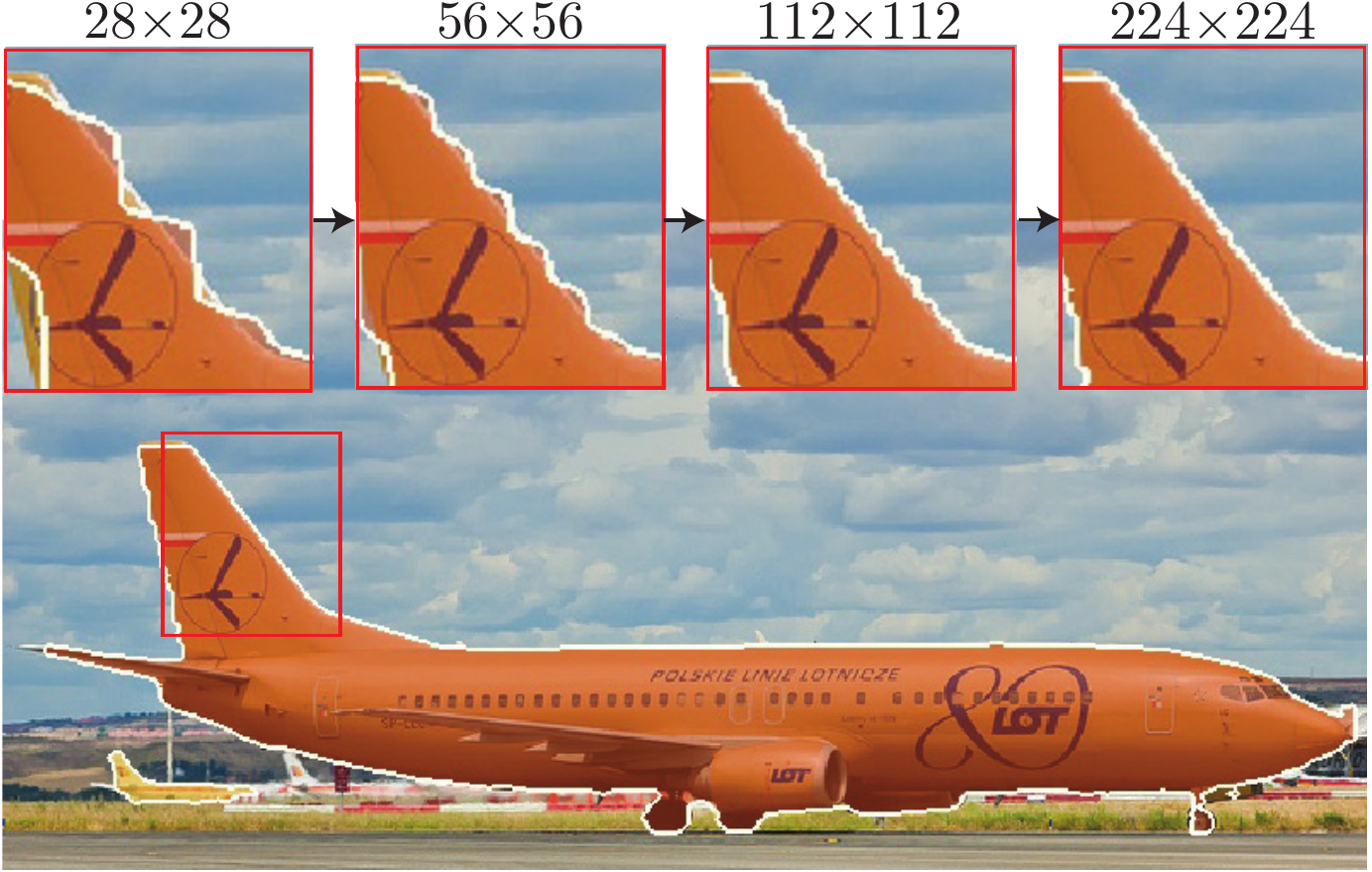}
  \caption{\bd{Anti-aliasing with \name.} Precise object delineation requires output mask resolution to match or exceed the resolution of the input image region that the object occupies.}
\label{fig:high-res}
\end{figure}

Table~\ref{tab:subdivision} shows \name subdivision inference with different output resolutions and number of points selected at each subdivision step. Predicting masks at a higher resolution can improve results. Though AP saturates, visual improvements are still apparent when moving from lower (\eg, 56$\x$56) to higher (\eg, 224$\x$224) resolution outputs, see Fig.~\ref{fig:high-res}. AP also saturates with the number of points sampled in each subdivision step because points are selected in the most ambiguous areas first. Additional points may make predictions in the areas where a coarse prediction is already sufficient. For objects with complex boundaries, however, using more points may be beneficial.

\subsection{Ablation Experiments}

We conduct a number of ablations to analyze \name. In general we note that it is robust to the exact design of the point head MLP. Changes of its depth or width do not show any significant difference in our experiments.

\begin{table}[t]\centering
\tablestyle{5pt}{1.0}\begin{tabular}{@{}l|cc|c@{}}
  & \multicolumn{2}{c|}{COCO} & Cityscapes \\
selection strategy & AP & \aplvis & AP \\\shline
regular grid             & 35.7 & 39.1 & 34.4 \\
uniform ($k{=}1, \beta{=}0.0$)         & 35.9 & 39.0 & 34.5 \\
mildly biased ($k{=}3, \beta{=}0.75$)   & \bd{36.3} & \bd{39.7} & \bd{35.8} \\
heavily biased ($k{=}10, \beta{=}1.0$)   & 34.4 & 37.5 & 34.1 \\
\end{tabular}
\vspace{-1mm}
\caption{\bd{Training-time point selection strategies} with \sqr{14} points per box. Mildly biasing sampling towards uncertain regions performs the best. Heavily biased sampling performs even worse than uniform or regular grid sampling indicating the importance of coverage. \aplvis is COCO mask AP evaluated against the higher-quality LVIS annotations~\cite{gupta2019lvis} (see text for details).
\label{tab:sampling}}
\end{table}

\paragraph{Point selection during training.} During training we select \sqr{14} points per object following the biased sampling strategy (\S\ref{subsec:sampling}). Sampling only \sqr{14} points makes training computationally and memory efficient and we found that using more points does not improve results. Surprisingly, sampling only 49 points per box still maintains AP, though we observe an increased variance in AP.

Table~\ref{tab:sampling} shows \name performance with different selection strategies during training. Regular grid selection achieves similar results to uniform sampling. Whereas biasing sampling toward ambiguous areas improves AP. However, a sampling strategy that is biased too heavily towards boundaries of the coarse prediction ($k{>}10$ and $\beta$ close to 1.0) decreases AP. Overall, we find a wide range of parameters $2{<}k{<}5$ and ${0.75}{<}\beta{<}1.0$ delivers similar results.

\begin{table}[t]\centering
\tablestyle{5pt}{1.0}\begin{tabular}{@{}l|c|cc@{}}
  & & \multicolumn{2}{c}{COCO} \\
mask head & backbone & AP & \aplvis \\\shline
4$\x$ conv & R50-FPN & 37.2 \phantom{+1.0} & 39.5 \phantom{+2.0}\\
PointRend  & R50-FPN & \bd{38.2 \dt{+1.0}} & \bd{41.5 \dt{+2.0}}\\ \hline
4$\x$ conv & R101-FPN & 38.6 \phantom{+1.2} & 41.4 \phantom{+2.1}\\
PointRend  & R101-FPN & \bd{39.8 \dt{+1.2}} & \bd{43.5 \dt{+2.1}}\\ \hline
4$\x$ conv & X101-FPN & 39.5 \phantom{+1.4} & 42.1 \phantom{+2.8}\\
PointRend  & X101-FPN & \bd{40.9 \dt{+1.4}} & \bd{44.9 \dt{+2.8}}\\
\end{tabular}
\caption{\bd{Larger models and a longer 3$\x$ schedule~\cite{he2019rethinking}.} \name benefits from more advanced models and the longer training. The gap between \name and the default mask head in Mask R-CNN holds. \aplvis is COCO mask AP evaluated against the higher-quality LVIS annotations~\cite{gupta2019lvis} (see text for details).
\label{tab:large-models}}
\end{table}

\paragraph{Larger models, longer training.} Training ResNet-50 + FPN (denoted R50-FPN) with the 1$\x$ schedule under-fits on COCO. In Table~\ref{tab:large-models} we show that the \name improvements over the baseline hold with both longer training schedule and larger models (see the appendix for details).
 
\begin{figure*}
\includegraphics[width=1.0\linewidth]{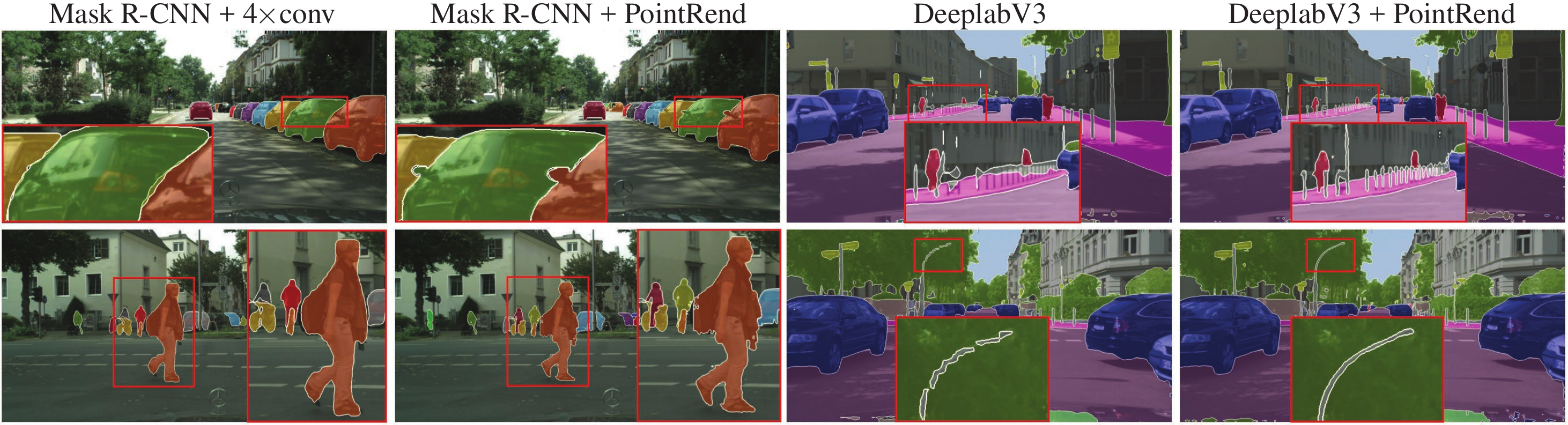}
  \caption{\textbf{Cityscapes example results for instance and semantic segmentation.} In instance segmentation larger objects benefit more from \name ability to yield high resolution output. Whereas for semantic segmentation \name recovers small objects and details.}
\vspace{-2mm}
\label{fig:examples-cs}
\end{figure*}

\section{Experiments: Semantic Segmentation}\label{sec:semantic}

\name is not limited to instance segmentation and can be extended to other pixel-level recognition tasks. Here, we demonstrate that \name can benefit two semantic segmentation models: DeeplabV3~\cite{deeplabV3}, which uses dilated convolutions to make prediction on a denser grid, and SemanticFPN~\cite{kirillov2019panopticfpn}, a simple encoder-decoder architecture.

\paragraph{Dataset.} We use the Cityscapes~\cite{Cordts2016Cityscapes} semantic segmentation set with 19 categories, 2975 training images, and 500 validation images. We report the median mIoU of 5 trials.

\paragraph{Implementation details.} We reimplemented DeeplabV3 and SemanticFPN following their respective papers. SemanticFPN uses a standard ResNet-101~\cite{he2016deep}, whereas DeeplabV3 uses the ResNet-103 proposed in~\cite{deeplabV3}.\footnote{It replaces the ResNet-101 res$_1$ 7$\x$7 convolution with three 3$\x$3 convolutions (hence ``ResNet-103'').} We follow the original papers' training schedules and data augmentation (details are in the appendix).

We use the same \name architecture as for instance segmentation. Coarse prediction features come from the (already coarse) output of the semantic segmentation model. Fine-grained features are interpolated from res$_2$ for DeeplabV3 and from P$_2$ for SemanticFPN. During training we sample as many points as there are on a stride 16 feature map of the input (2304 for deeplabV3 and 2048 for SemanticFPN). We use the same $k{=}3, \beta{=}0.75$ point selection strategy. During inference, subdivision uses $N{=}8096$ (\ie, the number of points in the stride 16 map of a 1024$\x$2048 image) until reaching the input image resolution. To measure prediction uncertainty we use the same strategy during training and inference: the difference between the most confident and second most confident class probabilities.

\begin{table}[t]\centering
\tablestyle{5pt}{1.0}\begin{tabular}{@{}l|c|l@{}}
method & output resolution & mIoU \\\shline
DeeplabV3-OS-16         & 64$\x$128 & 77.2 \\
DeeplabV3-OS-8          & 128$\x$256 & 77.8 \dt{+0.6}\\
DeeplabV3-OS-16 + \name & 1024$\x$2048 & \bd{78.4 \dt{+1.2}}  \\
\end{tabular}
\vspace{-1mm}
\caption{\bd{DeeplabV3 with \name} for Cityscapes semantic segmentation outperforms baseline DeepLabV3. Dilating the res$_4$ stage during inference yields a larger, more accurate prediction, but at much higher computational and memory costs; it is still inferior to using \name.
\label{tab:semseg-deeplabv3}}
\end{table}

\begin{figure}
\includegraphics[width=1.0\linewidth]{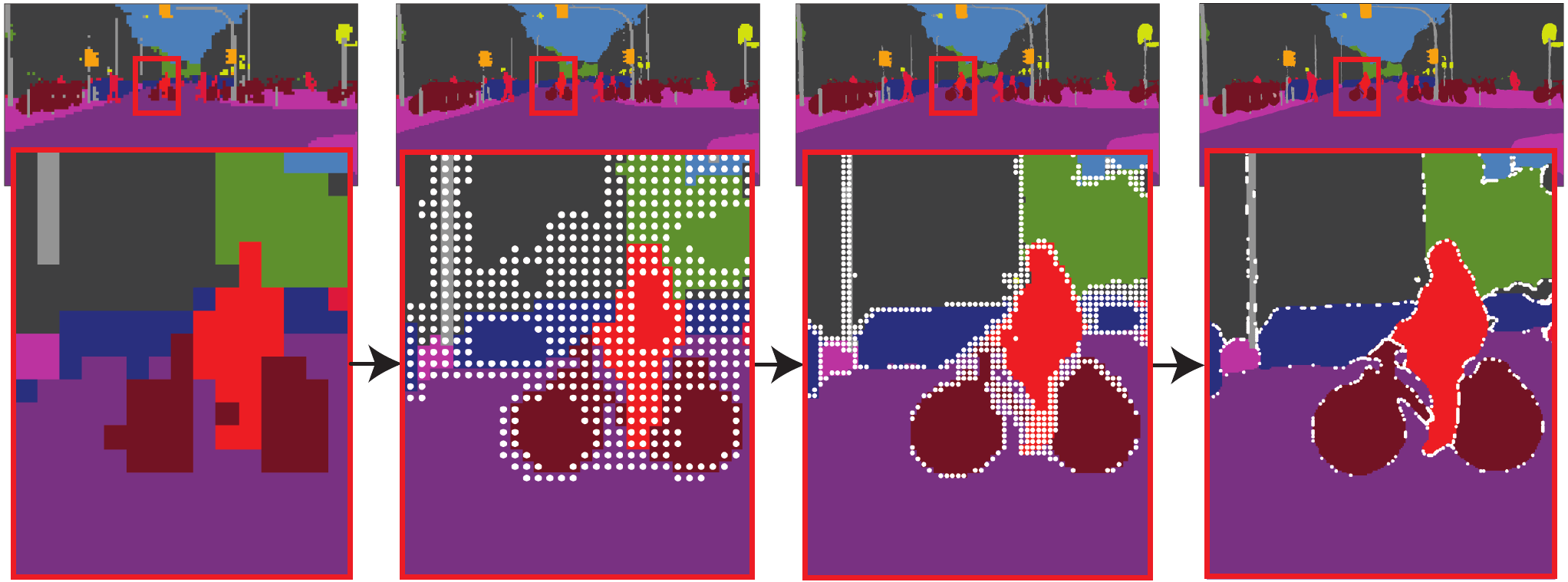}
  \caption{\textbf{\name inference for semantic segmentation.} \name refines prediction scores for areas where a coarser prediction is not sufficient. To visualize the scores at each step we take $\arg\max$ at given resolution without bilinear interpolation.
 }
\vspace{-2mm}
\label{fig:sem-inference}
\end{figure}

\paragraph{DeeplabV3.} In Table~\ref{tab:semseg-deeplabv3} we compare DeepLabV3 to DeeplabV3 with \name. The output resolution can also be increased by 2$\x$ at inference by using dilated convolutions in res$_4$ stage, as described in~\cite{deeplabV3}. Compared to both, \name has higher mIoU.
Qualitative improvements are also evident, see Fig.~\ref{fig:examples-cs}. By sampling points adaptively, \name reaches 1024$\x$2048 resolution (\ie 2M points) by making predictions for only 32k points, see Fig.~\ref{fig:sem-inference}.

\begin{table}[t]\centering
\tablestyle{4pt}{1.0}\begin{tabular}{@{}l|c|l@{}}
method & output resolution & mIoU \\\shline
SemanticFPN P$_2$-P$_5$         & 256$\x$512   & 77.7 \\
SemanticFPN P$_2$-P$_5$ + \name & 1024$\x$2048 & \bd{78.6 \dt{+0.9}}\\ \hline
SemanticFPN P$_3$-P$_5$         & 128$\x$256   & 77.4 \\
SemanticFPN P$_3$-P$_5$ + \name & 1024$\x$2048 & \bd{78.5 \dt{+1.1}}\\
\end{tabular}
\vspace{-1mm}
\caption{\bd{SemanticFPN with \name} for Cityscapes semantic segmentation outperform the baseline SemanticFPN.
\label{tab:semseg-semanticfpn}}
\end{table}

\paragraph{SemanticFPN.} Table~\ref{tab:semseg-semanticfpn} shows that SemanticFPN with \name improves over both 8$\x$ and 4$\x$ output stride variants without \name.

\begin{appendices}
\section{Instance Segmentation Details}

We use SGD with 0.9 momentum; a linear learning rate warmup~\cite{goyal2017accurate} over 1000 updates starting from a learning rate of 0.001 is applied; weight decay 0.0001 is applied; horizontal flipping and scale train-time data augmentation; the batch normalization (BN)~\cite{ioffe2015batch} layers from the ImageNet pre-trained models are frozen (\ie, BN is not used); no test-time augmentation is used.

\paragraph{COCO~\cite{lin2014coco}:} 16 images per mini-batch; the training schedule is 60k / 20k / 10k updates at learning rates of 0.02 / 0.002 / 0.0002 respectively; training images are resized randomly to a shorter edge from 640 to 800 pixels with a step of 32 pixels and inference images are resized to a shorter edge size of 800 pixels. 

\paragraph{Cityscapes~\cite{Cordts2016Cityscapes}:} 8 images per mini-batch the training schedule is 18k / 6k updates at learning rates of 0.01 / 0.001 respectively; training images are resized randomly to a shorter edge from 800 to 1024 pixels with a step of 32 pixels and inference images are resized to a shorter edge size of 1024 pixels.

\paragraph{Longer schedule:} The 3$\x$ schedule for COCO is 210k / 40k / 20k updates at learning rates of 0.02 / 0.002 / 0.0002, respectively; all other details are the same as the setting described above.

\section{Semantic Segmentation Details}

\paragraph{DeeplabV3~\cite{deeplabV3}:} We use SGD with 0.9 momentum with 16 images per mini-batch cropped to a fixed 768$\x$768 size; the training schedule is 90k updates with a poly learning rate~\cite{liu2015parsenet} update strategy, starting from 0.01; a linear learning rate warmup~\cite{goyal2017accurate} over 1000 updates starting from a learning rate of 0.001 is applied; the learning rate for ASPP and the prediction convolution are multiplied by 10; weight decay of 0.0001 is applied; random horizontal flipping and scaling of 0.5$\x$ to 2.0$\x$ with a 32 pixel step is used as training data augmentation; BN is applied to 16 images mini-batches; no test-time augmentation is used; 

\paragraph{SemanticFPN~\cite{kirillov2019panopticfpn}:} We use SGD with 0.9 momentum with 32 images per mini-batch cropped to a fixed 512$\x$1024 size; the training schedule is 40k / 15k / 10k updates at learning rates of 0.01 / 0.001 / 0.0001 respectively; a linear learning rate warmup~\cite{goyal2017accurate} over 1000 updates starting from a learning rate of 0.001 is applied; weight decay 0.0001 is applied; horizontal flipping, color augmentation~\cite{liu2016ssd}, and crop bootstrapping~\cite{bulo2017place} are used during training; scale train-time data augmentation resizes an input image from 0.5$\x$ to 2.0$\x$ with a 32 pixel step; BN layers are frozen (\ie, BN is not used); no test-time augmentation is used.

\section{\aplvis Computation}

The first version (v1) of this paper on arXiv has an error in COCO mask AP evaluated against the LVIS annotations~\cite{gupta2019lvis} (\aplvis). The old version used an incorrect list of the categories not present in each evaluation image, which resulted in lower \aplvis values.

\end{appendices}

{\small \bibliographystyle{ieee_fullname} \bibliography{point_rend}}

\end{document}